\documentclass{article}

\PassOptionsToPackage{numbers,sort&compress}{natbib}
\usepackage[final,sglblindworkshop]{neurips_2026}
\workshoptitle{World Models for High-Stakes Health. Selected for Oral Presentation}

\usepackage[utf8]{inputenc}
\usepackage[T1]{fontenc}
\usepackage{hyperref}
\usepackage{url}
\usepackage{booktabs}
\usepackage{amsmath,amsfonts,amssymb}
\usepackage{nicefrac}
\usepackage{microtype}
\usepackage{xcolor}
\usepackage{graphicx}
\usepackage{subcaption}
\usepackage{multirow}
\usepackage{array}
\usepackage{enumitem}
\usepackage{mathtools}
\usepackage{float}

\title{Repeated-Measure Leakage, Distribution Shift, and Reliability under Partial Observation in Patient World Models}
\author{%
Arjun Subramanian\\
Massachusetts Institute of Technology\\
Computer Science and Artificial Intelligence Laboratory (CSAIL)\\
\texttt{arjun\_s@mit.edu}
}

\hypersetup{
  pdftitle={Repeated-Measure Leakage, Distribution Shift, and Reliability under Partial Observation in Patient World Models},
  pdfauthor={Arjun Subramanian}
}

\newcommand{\nmse}{\mathrm{NMSE}}
\newcommand{\R}{\mathbb{R}}

\begin{document}
\maketitle

\begin{abstract}
Patient world models are increasingly proposed for longitudinal prediction, intervention-aware reasoning, and clinical-trial simulation. Causal or clinical intervention validity is distinct from predictive generalization and reliability; before making stronger claims, the underlying predictive state should generalize across patients, survive realistic shifts and missing observations, and expose failure through meaningful reliability signals. We evaluate these prerequisites in a deliberately narrow setting: short-horizon digital-biomarker forecasting from PhysioNet GaitPDB, comprising 165 participants, 306 recordings, and 51,129 context-future pairs. Using persistence, ridge, MLP, GRU, Transformer, and a compact JEPA-style predictor, we build an evaluation ladder that progressively removes raw temporal overlap, same-recording familiarity, and same-patient familiarity before testing unseen-patient generalization. For GRU, NMSE rises from 0.1227 under random-window splitting to 0.1393 after eliminating raw train-test overlap and to 0.1961 under patient holdout. Among 54 participants with repeated recordings, exposure to a different recording from the same patient improves GRU NMSE from 0.2177 to 0.1556, while a recording-excluded identity hypothesis is not supported at the participant level. Under participant-held-out evaluation, MLP and Transformer are statistically indistinguishable. Study shift, a four-times-longer prediction gap, and partial observation further degrade performance; under 50\% temporal masking, Transformer NMSE rises to 0.611 while MC-dropout predictive variance falls. We do not claim a longitudinal or intervention-aware simulator. Instead, the results support a prerequisite evaluation stack of patient separation, repeated-measure controls, shift, missingness, and uncertainty validation before stronger patient-world-model claims are trusted.
\end{abstract}

\section{Introduction}
\label{sec:intro}

World models learn predictive states intended to support forecasting, rollouts, and eventually decision making \citep{ha2018world}. In medicine, recent systems extend that ambition toward treatment-conditioned disease evolution, long-horizon patient trajectories, and intervention-aware simulation \citep{yang2025medical,mu2026ehrworld,wang2026chronomedicalworld}. Counterfactual or treatment-conditioned causal validity is conceptually distinct from predictive generalization and reliability. Before trusting a simulated intervention, one should first establish that the predictive state used to generate that simulation generalizes across patients, does not benefit spuriously from repeated-measure familiarity, remains usable under distribution shift and partial observation, and produces reliability signals that track failure. These tests are prerequisites for stronger intervention claims; they do not themselves establish causal or clinical validity.

We therefore ask a prerequisite question rather than claiming to build a longitudinal clinical simulator: \emph{what should a predictive patient-state model have to survive before stronger patient-world-model claims are credible?} Our testbed is deliberately narrow and falsifiable: short-horizon forecasting of vertical ground-reaction-force (VGRF) signals within gait recordings. It is modest in clinical scope, but it exposes three structural properties that recur in high-stakes longitudinal data: repeated measurements from individuals, heterogeneous acquisition protocols, and incomplete observations. This setting lets us isolate evaluation failure modes without conflating them with treatment-effect identification or clinical utility.

Repeated-measures leakage is already known to distort clinical machine-learning estimates. Record-wise validation can substantially overestimate generalization when the intended use is prediction for unseen people \citep{saeb2017usecase}; participant-aware analyses in biomechanics reach similar conclusions \citep{karbalaie2026participant}. Yet a random-window versus patient-held-out comparison still leaves a mechanistic ambiguity: adjacent train/test windows may share raw samples or near-duplicate waveform context, so an observed gap cannot by itself distinguish temporal overlap from persistent same-patient familiarity. We address that ambiguity with an \emph{evaluation ladder} that successively removes (i) raw temporal overlap, (ii) same-recording familiarity, and (iii) same-patient familiarity before reaching unseen-patient evaluation.

Our contribution is an evaluation methodology and empirical stress test, not a new JEPA method. The novelty is not participant-level splitting by itself, but decomposing evaluation optimism into distinct familiarity levels and then asking whether conclusions survive additional stresses that matter for patient-world-model reliability. Specifically, we:
\begin{itemize}[leftmargin=*,nosep]
    \item construct a four-level repeated-measure ladder spanning random-window, blocked/no-overlap, same-patient/new-recording, and participant-held-out evaluation, with programmatic zero-overlap checks where claimed;
    \item quantify persistent same-patient benefit using a matched cross-recording control on 54 participants, while separating descriptive identity structure from unsupported causal memorization claims;
    \item evaluate six predictors under leakage-resistant five-fold patient holdout with training-side model selection, train-only preprocessing, participant-level inference, and three seeds for stochastic models;
    \item test whether predictive-state performance and reliability survive internal study shift, a four-times-longer prediction gap, partial observation, and lightweight MC-dropout evaluation without adapting on held-out patients.
\end{itemize}

\textbf{Scope of claim.} This is not a model leaderboard and not a clinical-trial simulator. MLP and Transformer are tied under participant-held-out evaluation, the compact JEPA comparator is worse in signal-space forecasting, and several favorable conclusions weaken under stricter evaluation. We interpret these results as a prerequisite evaluation layer: if predictive states fail patient separation, shift, missingness, or reliability stress, downstream treatment-conditioned or counterfactual rollouts should not receive stronger trust merely because they are more ``world-model-like.'' These tests concern predictive generalization and reliability, not clinical or causal validity, and the observed effect sizes are not claimed to transport unchanged to other diseases or datasets. The gait-forecasting task is therefore used as a stress-test substrate for the evaluation methodology, not as a claim that short-horizon gait forecasting itself constitutes a longitudinal patient world model.

\section{Related Work}
\label{sec:related}

\paragraph{Patient world models and prerequisite evaluation.}
Classical world models learn compact predictive dynamics for downstream control or planning \citep{ha2018world}. Recent medical systems instantiate this idea with treatment-conditioned tumor evolution \citep{yang2025medical}, long-horizon EHR trajectories \citep{mu2026ehrworld}, and action-conditioned chronic-care models \citep{wang2026chronomedicalworld}. Such systems ultimately face at least two distinct validation questions: whether their predictive state transfers reliably, and whether intervention-conditioned rollouts have causal or clinical validity. Our setting addresses only the first. It contains no treatment, intervention, disease-progression, or causal variable; instead, it tests whether short-horizon predictive states satisfy generalization and reliability conditions that stronger patient-world-model claims presuppose.

\paragraph{Repeated measures and validation.}
Clinical sensor datasets often contain many correlated samples per person. \citet{saeb2017usecase} showed that record-wise validation can dramatically overstate performance relative to subject-wise validation, and participant-aware cross-validation remains an active concern in digital health and biomechanics \citep{karbalaie2026participant}. Small effective sample sizes further motivate uncertainty over participants rather than over repeated windows \citep{varoquaux2018cv}. We preserve the complete participant--recording--window hierarchy and add intermediate controls that distinguish raw temporal overlap from persistent same-patient familiarity.

\paragraph{Predictive representation learning and uncertainty.}
Joint-embedding predictive architectures learn by predicting target representations instead of reconstructing observations \citep{assran2023ijepa}; related work adapts this idea to time series \citep{ennadir2025tsjepa}. We include a compact JEPA-style model only as a comparator. For reliability, MC dropout offers a lightweight predictive-disagreement heuristic \citep{gal2016dropout}, while selective prediction studies error as uncertain examples are rejected \citep{geifman2017selective}. Predictive uncertainty can fail under distribution shift \citep{ovadia2019uncertainty}; we test whether it also behaves sensibly under controlled missingness.

\section{Evaluation Design}
\label{sec:methods}

\subsection{Dataset hierarchy and predictive task}
We use PhysioNet's \emph{Gait in Parkinson's Disease} database (GaitPDB) v1.0.0 \citep{hausdorff2008gaitpdb,pollard2026physionet}. The public description reports 93 participants with Parkinson's disease (PD) and 73 healthy controls (HC). A file-level audit of recording names and \texttt{demographics.txt} instead finds 93 valid PD IDs and 72 valid HC IDs: \textbf{165 usable participants}, \textbf{306 recordings}, and no corrupt 19-column VGRF files. We retain this file-authoritative cohort and document the one-count discrepancy without assuming that a malformed metadata row explains it (Appendix~\ref{app:data}).

Each recording contains 18 VGRF channels at 100~Hz. For multivariate signal $x\in\R^{T\times18}$ we construct
\begin{align}
 c_t &= x[t-128:t], & y_t &= x[t+16:t+48],
\end{align}
with stride 64 samples: 1.28~s context, 0.16~s prediction gap, and 0.32~s target. The corpus yields 51,129 windows. This is \emph{within-recording digital-biomarker forecasting}, not longitudinal disease progression. Fold-specific normalization is fit on training data only. The primary signal metric is
\begin{equation}
\nmse(\hat y,y)=\frac{\|\hat y-y\|_2^2/|y|}{\sigma^2_{\mathrm{train}}},
\end{equation}
with MAE and Pearson correlation as complementary metrics.

\subsection{An evaluation ladder for repeated measurements}
\label{sec:ladder}
Our primary unseen-patient protocol is \textbf{participant-held-out}: stratified five-fold outer cross-validation keeps every recording and window from a participant in one fold. Outer-test participants never enter normalization, architecture selection, early stopping, or calibration.

We then use three diagnostic protocols to localize why easier evaluation looks better. \textbf{Random-window-leaky} randomly divides windows so all 165 participants appear in both train and test; adjacent windows can share raw temporal structure. \textbf{Blocked/no-overlap} still permits the same recording and participant on both sides, but partitions recordings into contiguous blocks with a 352-sample safety gap and programmatically verifies that no context, prediction-gap, or target raw index overlaps across train/validation/test. Finally, \textbf{cross-recording same-patient} uses the 54 participants with at least two recordings: the test recording is absent from training, validation, and scaler fitting, while another recording from the same participant may be seen. This last control tests persistent patient familiarity without exact-recording familiarity. None of these diagnostic protocols is presented as clinically valid generalization.

\begin{table}[t]
\centering
\caption{Evaluation ladder. ``Yes'' means that source of familiarity may cross the train--test boundary by design. Only the final row targets unseen-patient generalization.}
\label{tab:ladder}
\small
\setlength{\tabcolsep}{4.0pt}
\begin{tabular}{lcccc}
\toprule
Protocol & Raw overlap & Same recording & Same patient & Unseen patient? \\
\midrule
Random-window & yes & yes & yes & no \\
Blocked/no-overlap & no & yes & yes & no \\
Same-patient/new-recording & no & no & yes & no \\
Participant-held-out & no & no & no & yes \\
\bottomrule
\end{tabular}
\end{table}

\subsection{Models, selection, seeds, and inference}
We evaluate persistence; Ridge autoregression ($\alpha=10$); a two-layer ReLU MLP (hidden 128); a two-layer GRU (hidden 128, 291k parameters); a two-layer Transformer (width 64, four heads, FFN 128, 110k parameters); and a compact JEPA-style predictor (151k trainable parameters, latent 128). JEPA uses a context encoder, EMA target encoder ($\tau=0.996$), latent predictor, latent MSE, and a small variance hinge; it contains no action, treatment, counterfactual, intervention, or physical-grounding module. Signal-space JEPA metrics use a training-only Ridge readout from predicted latents; latent cosine is kept separate.

GRU and Transformer each use two fixed candidate architectures. Within every outer fold, selection occurs on an inner participant split of outer-training data only; all five folds select the same smaller candidate. That configuration is frozen across seeds. The primary participant-held-out MLP, GRU, Transformer, and JEPA experiments use seeds 0, 1, and 2; deterministic baselines and diagnostic protocols use one canonical seed. Neural models use AdamW, learning rate $10^{-3}$ for selected configurations, weight decay $10^{-4}$, batch size 128, gradient clipping at 1, cosine scheduling, at most 20 epochs, and training-side early stopping (Appendix~\ref{app:models}).

The inferential unit is always the participant. Where applicable, metrics aggregate windows $\rightarrow$ recordings $\rightarrow$ participants before testing. This targets performance for an unseen participant: after hierarchical aggregation, each participant receives equal inferential weight. Weighting by window count would instead give disproportionate influence to participants with longer or more numerous recordings and would answer a different, window-weighted estimand. We report 10,000-draw participant-bootstrap 95\% CIs, paired Wilcoxon signed-rank tests, paired Cohen's $d_z$, and Holm correction within analysis families; a global Holm column is retained in the artifact. We distinguish primary H1--H4, secondary H5--H6, and exploratory subgroup analyses; these hypotheses were not preregistered.
\paragraph{Reproducibility artifact.} Code, fixed manifests, configurations, analysis scripts, and paper-facing result tables are available through the Open Science Framework (OSF) at \href{https://doi.org/10.17605/OSF.IO/AUZVR}{https://doi.org/10.17605/OSF.IO/AUZVR}. Readers can reproduce the reported aggregate results from the saved outputs without retraining the neural models.

\subsection{Shift, partial observation, and reliability}
The three GaitPDB source-study prefixes (Ga, Ju, Si) differ in protocol and repeated-recording structure. \textbf{Leave-one-study-out} (LOSO) trains on two and tests on the third, with normalization and validation restricted to training studies. This is an internal combined study/acquisition/population shift stress test, not external validation. A separate \textbf{longer-horizon} control increases the prediction gap from 0.16 to 0.64~s while fixing context, target, folds, and architectures.

For \textbf{partial observation}, held-out contexts receive contiguous temporal masks of 25/50/75\%, channel removal of 25/50/75\%, or Gaussian noise at 0.25$\times$/1.0$\times$ training-channel standard deviation. Missing entries use training-side channel means; perturbation seeds are matched across models and targets stay clean. For \textbf{reliability}, eight MC-dropout passes define mean per-element predictive variance. Abstention thresholds and high-error cutoffs are chosen on training-side validation only. Reliability uses up to 32 evenly spaced windows per participant, independent of error; Transformer stochastic passes run on CPU because stochastic scaled-dot-product attention was unsupported on MPS in the tested environment.

\section{Results}
\label{sec:results}

\subsection{Participant holdout changes the model story}
Table~\ref{tab:main} gives the primary three-seed participant-held-out results. MLP and Transformer are statistically indistinguishable: NMSE 0.1769 and 0.1766, paired Holm $p=0.291$, with a 95\% CI on MLP$-$Transformer NMSE of $[-0.0032,0.0041]$. GRU is worse (0.1982), and the compact JEPA comparator is worse in signal-space forecasting (0.3101). Across-seed SD is small relative to participant uncertainty: 0.0013 (MLP), 0.0021 (GRU), 0.0009 (Transformer), and 0.0052 (JEPA). The same qualitative ordering holds descriptively within HC and PD strata (Appendix~\ref{app:subgroups}); we do not interpret between-diagnosis error differences clinically.

\begin{table}[t]
\centering
\caption{Primary participant-held-out performance ($n=165$ participants). CIs are participant-bootstrap 95\% intervals after window$\to$recording$\to$participant aggregation. Stochastic models average three seeds.}
\label{tab:main}
\small
\setlength{\tabcolsep}{4.1pt}
\begin{tabular}{lcccc}
\toprule
Model & NMSE (95\% CI) & MAE & Pearson & seeds \\
\midrule
Persistence & 2.9321 [2.7983, 3.0728] & 1.3519 & $-0.452$ & 1 \\
Ridge       & 0.2460 [0.2251, 0.2701] & 0.3219 & 0.878 & 1 \\
MLP         & \textbf{0.1769} [0.1660, 0.1888] & 0.2619 & 0.905 & 3 \\
GRU         & 0.1982 [0.1850, 0.2123] & 0.2594 & 0.897 & 3 \\
Transformer & \textbf{0.1766} [0.1666, 0.1875] & \textbf{0.2529} & \textbf{0.906} & 3 \\
JEPA        & 0.3101 [0.2916, 0.3302] & 0.3562 & 0.841 & 3 \\
\bottomrule
\end{tabular}
\end{table}

\subsection{Removing temporal overlap explains only part of evaluation inflation}
\label{sec:leakage-results}
\begin{figure}[t]
\centering
\includegraphics[width=\linewidth]{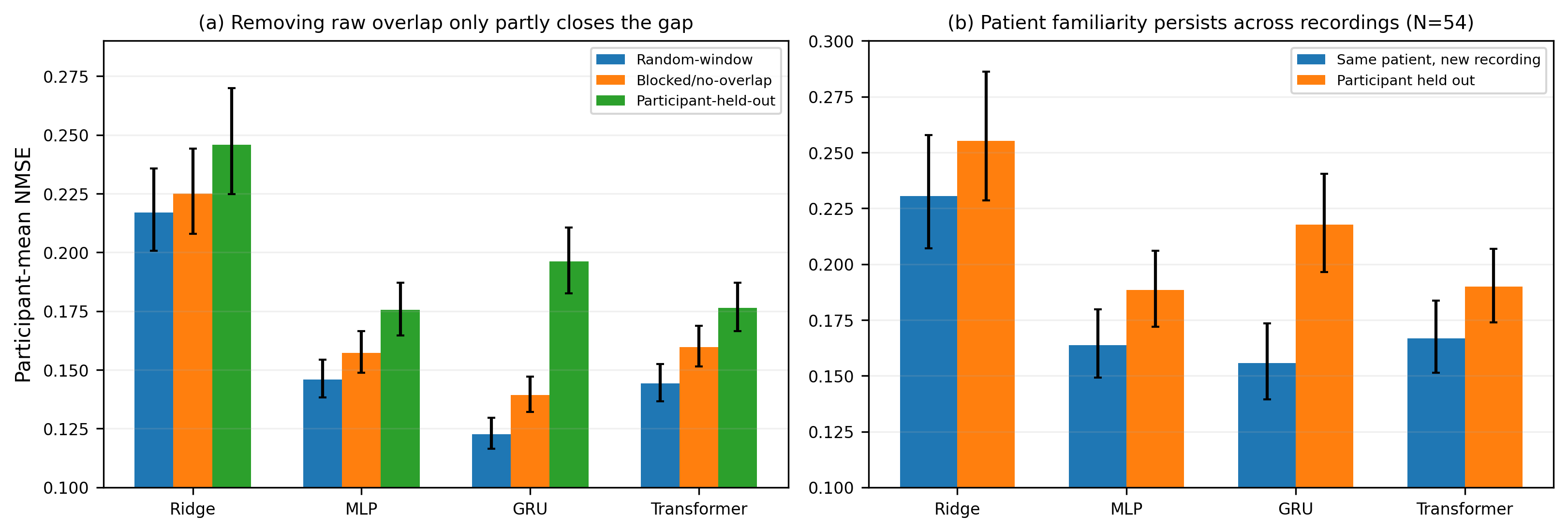}
\caption{Decomposing repeated-measure familiarity with seed-0 diagnostics. \textbf{Left:} random-window splitting is most optimistic; a blocked protocol removes all raw train--test sample overlap yet still permits the same recording and patient; patient holdout is strictest. \textbf{Right:} on the 54 participants with repeated recordings, training on a different recording from the same patient improves learned-model forecasting relative to matched patient holdout. Persistence is omitted from the panels for scale; it is essentially invariant in the left comparison and slightly worse under same-patient cross-recording evaluation. Error bars are participant-bootstrap 95\% CIs.}
\label{fig:leakage-controls}
\end{figure}

The seed-0 diagnostic ladder in Figure~\ref{fig:leakage-controls} separates increasingly strict forms of familiarity. Under random-window splitting, GRU NMSE is 0.1227. Eliminating every raw train--test overlap within the same recordings raises it to 0.1393, showing that temporal near-duplication contributes to the optimistic score. Yet blocked/no-overlap remains substantially better than true patient holdout (0.1961; secondary H5 mean $\Delta=-0.0568$, Holm $p=2.9\times10^{-24}$). The same pattern holds for MLP (0.1460 $\rightarrow$ 0.1572 $\rightarrow$ 0.1755), Transformer (0.1442 $\rightarrow$ 0.1598 $\rightarrow$ 0.1763), and Ridge (0.2170 $\rightarrow$ 0.2250 $\rightarrow$ 0.2458); blocked versus held-out is significant for all four learned models after Holm correction. Persistence is essentially unchanged across the three protocols (2.9220/2.9213/2.9319; H5 $p=0.514$), providing a negative control.

The cross-recording control asks a stricter question: does familiarity with the \emph{person} help when the entire test recording is unseen? Among the 54 eligible participants, GRU improves from matched participant-held-out NMSE 0.2177 to 0.1556 when a different recording from that same participant is available in training (H6 $\Delta=-0.0620$, Holm $p=1.9\times10^{-6}$). MLP improves by 0.0247 ($p=3.0\times10^{-4}$) and Transformer by 0.0232 ($p=2.0\times10^{-4}$). Persistence instead becomes slightly worse ($\Delta=+0.0408$). Thus exact temporal overlap explains part, but not all, of the optimistic random-window result; persistent patient-specific regularities remain a plausible contributor even after recording separation. Because this diagnostic deliberately exposes patient identity during training and has $N=54$, it is not an estimate of unseen-patient performance.

\subsection{Identity diagnostics are descriptive, and the confirmatory identity hypothesis fails}
Recording-excluded nearest-neighbor analysis gives high descriptive same-patient hit@1 relative to a nominal fold-level chance of 0.0086: GRU 0.209, Transformer 0.189, MLP 0.156, and JEPA 0.118. Same-patient/different-recording cosine similarity also exceeds between-patient similarity for GRU and Transformer (Appendix~\ref{app:identity}). However, the participant-level H2 tests against the nominal chance reference are \emph{not} significant after correction ($p_{\mathrm{Holm}}=1$ for all four neural representation models). This is an important negative result: query-level retrieval structure should not be promoted to confirmatory evidence of patient memorization. Persistence further cautions against a mechanistic interpretation: it attains recording-excluded hit@1 of 0.143 despite showing almost no split-induced NMSE inflation. We therefore treat identity structure as descriptive context, while the cross-recording forecasting control provides the cleaner performance-based evidence that patient familiarity can matter.

\subsection{Study shift and a longer prediction gap weaken generalization}
\begin{figure}[t]
\centering
\includegraphics[width=\linewidth]{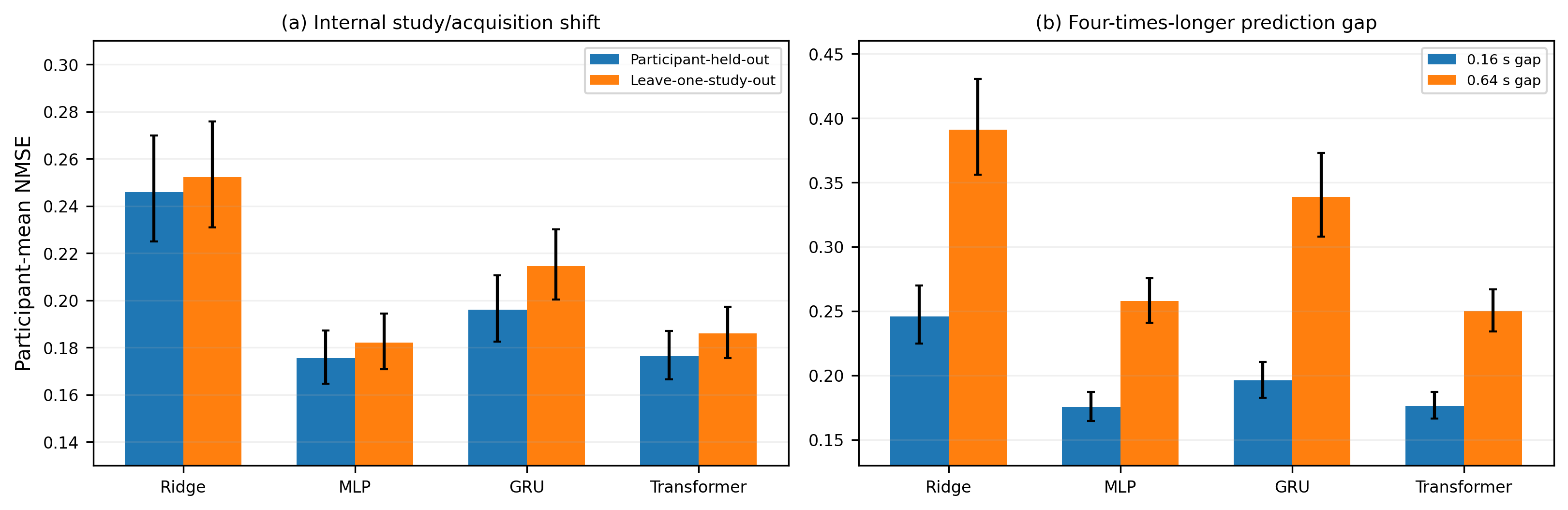}
\caption{Generalization stress tests for learned predictors. \textbf{Left:} LOSO is internal to GaitPDB and combines acquisition/population shift; it is not external validation. \textbf{Right:} increasing the prediction gap from 0.16 to 0.64~s worsens NMSE under unchanged participant folds and frozen architectures. Persistence is omitted for scale and discussed in text. Error bars are participant-bootstrap 95\% CIs.}
\label{fig:shift}
\end{figure}

LOSO testing modestly but consistently worsens learned-model NMSE relative to ordinary participant-held-out CV (Figure~\ref{fig:shift}). MLP rises from 0.1756 to 0.1820 (H4 paired $\Delta=0.0064$, Holm $p=1.5\times10^{-4}$), GRU from 0.1962 to 0.2146 ($\Delta=0.0184$, $p=3.0\times10^{-8}$), and Transformer from 0.1764 to 0.1859 ($\Delta=0.0096$, $p=8.7\times10^{-9}$). Ga, Ju, and Si differ not only in acquisition protocol but also in age distribution and repeated-recording structure (Appendix~\ref{app:data}); LOSO therefore measures a combined internal shift, not external validation.

Increasing the prediction gap from 0.16 to 0.64~s produces a larger degradation with frozen models: MLP rises to 0.2577 ($\Delta=0.0821$), Transformer to 0.2497 ($\Delta=0.0734$), and GRU to 0.3385 ($\Delta=0.1424$), all Holm $p<4\times10^{-28}$. Persistence improves from 2.93 to 1.07 because a longer offset changes the relation between the last observed gait state and the periodic target; this task-specific reversal reinforces why baseline behavior should be interpreted mechanistically rather than as a generic ranking. The 0.64~s experiment remains short-horizon waveform forecasting, not disease-course prediction.

\subsection{Missingness can break prediction while uncertainty decreases}
\begin{figure}[t]
\centering
\includegraphics[width=\linewidth]{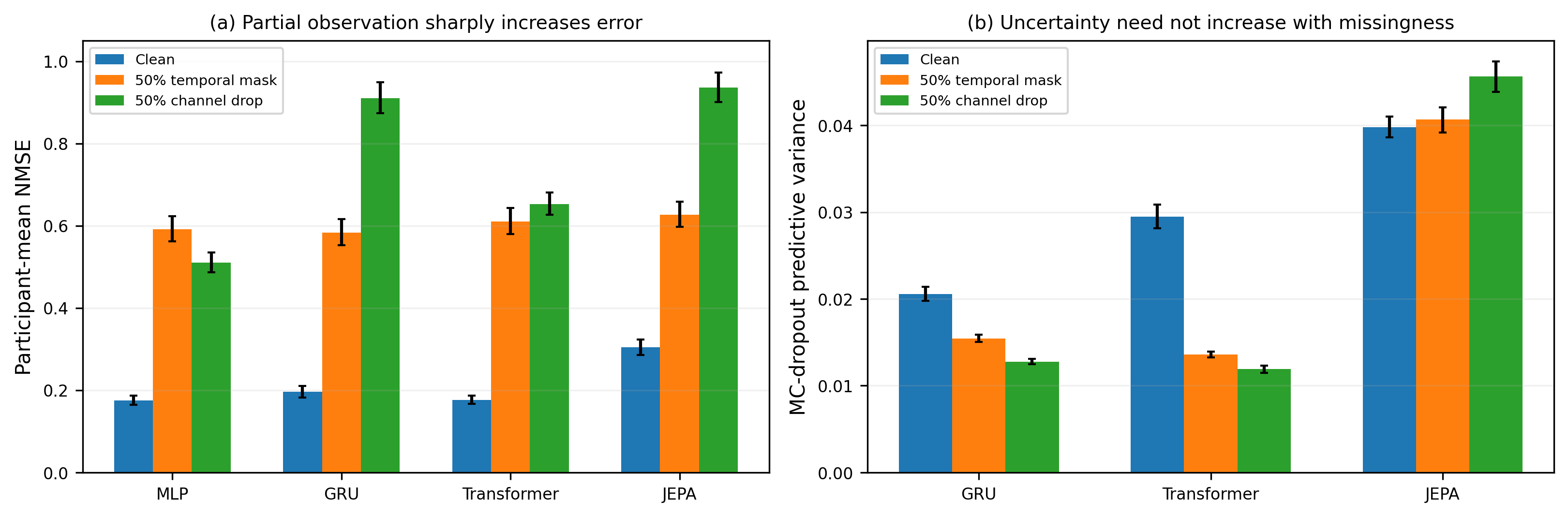}
\caption{Missingness exposes a reliability mismatch. \textbf{Left:} 50\% temporal masking and channel removal sharply increase participant-level NMSE. \textbf{Right:} GRU and Transformer MC-dropout predictive variance decreases under the same mean-filled corruptions, whereas JEPA's weaker dropout mechanism moves differently. Corruptions use training-side statistics and are applied only at evaluation. Error bars are participant-bootstrap 95\% CIs.}
\label{fig:robustness}
\end{figure}

Partial observation creates the largest learned-model degradation (Figure~\ref{fig:robustness}, left). Under 50\% temporal masking, seed-0 NMSE rises from 0.1754 to 0.5918 for MLP, 0.1961 to 0.5835 for GRU, and 0.1763 to 0.6106 for Transformer. Under 50\% channel removal, the same models reach 0.5102, 0.9102, and 0.6530. All six primary clean-versus-50\%-corruption H3 comparisons have Holm $p=4.75\times10^{-28}$, with $d_z=2.55$--3.79. Ridge is particularly brittle to additive noise (medium-noise NMSE 6.13).

The uncertainty heuristic does not track this deterioration. On the reliability subset, Transformer NMSE changes from 0.1788 clean to 0.6079 under 50\% temporal masking and 0.6523 under 50\% channel removal, while MC-dropout predictive variance falls from 0.0295 to 0.0136 and 0.0119, respectively; at 75\% channel removal it falls to 0.0081. On clean data, participant-mean within-person Spearman correlation between uncertainty and error is only 0.115 for GRU and 0.079 for Transformer, with high-error AUROC 0.615 and 0.575. Validation-chosen abstention thresholds provide little rescue: near 50\% target coverage, GRU improves from NMSE 0.200 to 0.183, whereas Transformer slightly worsens from 0.179 to 0.182 and JEPA is essentially unchanged. These are failures of the evaluated MC-dropout heuristic, not evidence of calibrated confidence.

\section{Discussion}
\label{sec:discussion}

\paragraph{Predictive generalization and reliability should precede intervention-validity claims.}
The central implication is methodological. Predictive generalization, reliability, and causal or clinical intervention validity are distinct evaluation targets. Patient world models may ultimately be judged by treatment-conditioned trajectories, counterfactual rollouts, or trial-level decisions, but a model that benefits from patient familiarity, degrades under acquisition shift or missing observations, or emits an uncertainty signal that moves opposite to its error has not yet established the predictive foundation needed to interpret those stronger evaluations. Our experiments therefore target a lower rung of the evaluation stack: not whether an intervention is causally correct, but whether the predictive state is sufficiently leakage-resistant, generalizable, and reliable to justify asking that stronger question next.

\paragraph{Leakage is hierarchical, not binary.}
The new diagnostic ladder changes the interpretation of the original random-window gap. Removing all raw overlap within a recording consistently raises learned-model error, so temporal near-duplication accounts for a real portion of the optimism. Yet blocked/no-overlap performance remains better than patient holdout, and the cross-recording control shows that a learned model can benefit from seeing the same patient even when the test recording is completely unseen. This is stronger evidence for persistent patient familiarity than same-recording nearest-neighbor retrieval, while still stopping short of a causal attribution: recording protocols, stable biomechanics, and patient-specific physiology are inseparable in this dataset.

\paragraph{Representation identity is not equivalent to leakage mechanism.}
The recording-excluded retrieval analysis is deliberately retained even though its participant-level H2 fails after correction. That null result, together with the persistence counterexample, prevents a tempting but unsupported story in which ``identity encoding'' directly causes every evaluation gap. The appropriate claim is narrower: learned models exploit information available when patient boundaries are relaxed, and some patient-identifiable structure survives recording exclusion, but the mechanism is not uniquely identified.

\paragraph{Architectural sophistication does not substitute for evaluation.}
The MLP and Transformer are tied under participant-held-out evaluation, while the compact JEPA-style model is worse in signal-space prediction. This is not evidence against JEPA as a general paradigm: the comparator is intentionally small and not equally architecture-searched. It does show that a model being more ``world-model-like'' does not, by itself, establish stronger cross-patient generalization. The evaluation protocol changes measured performance more than the distinction between the best conventional architectures here.

\paragraph{Patient holdout is necessary but insufficient.}
LOSO, longer-horizon, and missingness stress tests probe different failure axes after patient separation. The first reveals sensitivity to acquisition/population shift inside a single database; the second shows that immediate continuation is materially easier than a larger temporal gap; the third produces much larger failures under absent observations. Patient separation should therefore be treated as the beginning of reliable world-model evaluation rather than its endpoint. Real clinical trajectories are heterogeneous, sparse, and partially observed, so treatment-conditioned simulation built on a state that has not survived these stresses would conflate intervention validity with a more basic state-generalization failure.

\paragraph{Uncertainty must be evaluated as a prediction, too.}
MC dropout is attractive because it is lightweight and has a Bayesian interpretation under assumptions \citep{gal2016dropout}, but its predictive disagreement moves in the wrong direction under mean-filled missingness. This complements broader evidence that uncertainty estimates can fail under shift \citep{ovadia2019uncertainty}. A high-stakes model should therefore not receive credit for ``knowing when it is wrong'' merely because it emits an uncertainty score; the score itself needs corruption- and shift-aware validation.

\section{Limitations}
\label{sec:limitations}
GaitPDB is a single public resource; its three source studies are not independent external cohorts and differ in protocol, demographics, and repeated-recording structure. The specific effect sizes reported here should therefore not be assumed to transport to other diseases, modalities, institutions, or datasets; establishing such transportability requires independent external validation. The primary task predicts sub-second VGRF segments within a session, not longitudinal disease progression, treatment response, interventions, or causal effects. Blocked/no-overlap still exposes the same recording and participant, while cross-recording evaluation intentionally exposes the same participant and is limited to 54 people with repeats; both are diagnostic rather than clinically valid protocols. The recording-excluded identity hypothesis is not supported at the participant level after correction, so identity results remain descriptive. The 0.64~s control is still short-horizon forecasting. Partial-observation corruptions are synthetic rather than empirical deployment missingness. Reliability uses lightweight MC dropout instead of deep ensembles or a comprehensive uncertainty benchmark. JEPA was not run on the two new diagnostic splits and did not receive the same architecture search as GRU/Transformer. Finally, Apple MPS execution is reproducible at the protocol/configuration level but not guaranteed bitwise deterministic.

\section{Responsible-Use Statement}
\label{sec:responsible}
This work is secondary analysis of an existing public, de-identified PhysioNet dataset and introduces no new participant interaction. The models are research predictors for short-horizon gait signals; they are \textbf{not} diagnostic systems, treatment recommenders, clinical-trial simulators, or validated patient digital twins. The central potential benefit is methodological: identifying evaluation protocols that can overstate transferability and reliability before those errors propagate into higher-stakes claims. The corresponding risk is misuse of the reported numbers as evidence of clinical utility or of MC-dropout variance as calibrated confidence. We mitigate these risks by making patient holdout the primary protocol, fitting preprocessing/selection/calibration on training-side data only, using participant-level inference, retaining null and negative results, labeling same-patient and LOSO analyses as diagnostic/internal, and explicitly separating short-horizon forecasting from intervention-aware simulation. Any clinical use would require independent external validation, prospective assessment, clinically realistic missing-data studies, subgroup/fairness analysis, target-site uncertainty calibration, and---for intervention-aware claims---longitudinal treatment data with appropriate causal assumptions. No result here should be used to make decisions about an individual patient.

\section{Conclusion}
\label{sec:conclusion}
Before asking whether a patient world model can support treatment-conditioned rollouts, counterfactual reasoning, or clinical-trial simulation, one should first establish that its predictive state survives more basic tests of leakage-resistant generalization, robustness, and predictive reliability. In this digital-biomarker benchmark, raw temporal overlap explains only part of leakage-prone optimism; same-patient familiarity remains useful across distinct recordings; participant-held-out evaluation changes model comparisons; internal study shift, longer horizons, and missingness reduce performance; and a common uncertainty heuristic can become less responsive as prediction worsens. We therefore advocate an evaluation ladder---overlap control, recording control, patient separation, shift, missingness, and uncertainty validation---as a prerequisite layer before stronger longitudinal, causal, or intervention-aware claims are trusted.

\appendix

\section{Dataset Audit, Cohort Reconciliation, and Study Composition}
\label{app:data}
The public GaitPDB page describes 93 PD participants and 73 HC participants. Distributed recording filenames and valid participant rows in \texttt{demographics.txt} agree on 93 PD and 72 HC IDs. A malformed row labeled \texttt{Juc010} does not match the dataset participant-ID convention or a recording file; we do \emph{not} claim that it identifies the missing public-page control. All experiments therefore use 165 file-authoritative participants (72 HC, 93 PD), 306 recordings, and 51,129 primary windows. No corrupt 19-column VGRF recording was found.

\begin{table}[h]
\centering
\caption{File-authoritative hierarchy and source-study composition. Windows are context--target pairs, not participants.}
\small
\begin{tabular}{lrrrrr}
\toprule
Group & Participants & HC & PD & Recordings & Windows \\
\midrule
All & 165 & 72 & 93 & 306 & 51,129 \\
Ga (Yogev) & 47 & 18 & 29 & 113 & 21,005 \\
Ju (Hausdorff) & 54 & 25 & 29 & 129 & 18,156 \\
Si (Frenkel-Toledo) & 64 & 29 & 35 & 64 & 11,968 \\
\bottomrule
\end{tabular}
\end{table}

The source studies differ in repeated-recording structure: Ga has 1--3 recordings/person, Ju 1--7, and Si exactly 1. Mean recording duration is 120.5, 91.5, and 121.2~s, respectively. Mean age is 71.3, 66.0, and 59.9 years. These descriptive differences are why LOSO is interpreted as a combined study/acquisition/population shift rather than an isolated scanner/protocol effect or external validation.

\section{Diagnostic Split Construction and Hypothesis Families}
\label{app:splits}
For blocked/no-overlap evaluation, the occupied raw interval of each window includes context, prediction gap, and target. Recording blocks are separated by a 352-sample safety gap, twice the 176-sample full context--future span. Validators assert zero overlap of occupied raw indices between train/validation/test windows from the same recording. The split intentionally retains same-recording and same-patient exposure.

For cross-recording same-patient evaluation, 54 participants with at least two recordings (195 recordings total) are eligible. Each test recording is excluded from train, validation, and scaler fitting; a different recording from the same participant may appear in training. The matched participant-held-out comparison uses the same 54 IDs.

Primary families are H1 (random-window vs participant-held-out), H2 (recording-excluded identity vs nominal chance), H3 (50\% temporal/channel corruption vs clean), and H4 (LOSO vs ordinary held-out). Secondary H5 compares blocked/no-overlap with held-out, H6 compares cross-recording same-patient with held-out on eligible participants, and the 0.64~s horizon is a secondary control. HC/PD stratification is exploratory. These labels organize the final analysis but are not a preregistration.

\begin{table}[h]
\centering
\caption{Secondary repeated-measure controls. $\Delta$ is diagnostic-protocol NMSE minus matched participant-held-out NMSE; negative values mean the diagnostic protocol is easier. CIs and Holm $p$ values use participants as the unit.}
\small
\setlength{\tabcolsep}{3.8pt}
\begin{tabular}{lccrc}
\toprule
& \multicolumn{2}{c}{Blocked/no-overlap (H5, $N=165$)} & \multicolumn{2}{c}{Same-patient/new-recording (H6, $N=54$)} \\
Model & $\Delta$ [95\% CI] & Holm $p$ & $\Delta$ [95\% CI] & Holm $p$ \\
\midrule
Persistence & $-0.0106$ [$-0.0240$, 0.0024] & 0.514 & $+0.0408$ [0.0149,0.0651] & $7.4\times10^{-4}$ \\
Ridge & $-0.0208$ [$-0.0286$,$-0.0139$] & $1.1\times10^{-8}$ & $-0.0248$ [$-0.0433$,$-0.0090$] & 0.013 \\
MLP & $-0.0182$ [$-0.0243$,$-0.0129$] & $2.4\times10^{-9}$ & $-0.0247$ [$-0.0372$,$-0.0137$] & $3.0\times10^{-4}$ \\
GRU & $-0.0568$ [$-0.0671$,$-0.0473$] & $2.9\times10^{-24}$ & $-0.0620$ [$-0.0843$,$-0.0425$] & $1.9\times10^{-6}$ \\
Transformer & $-0.0166$ [$-0.0215$,$-0.0120$] & $1.7\times10^{-9}$ & $-0.0232$ [$-0.0343$,$-0.0132$] & $2.0\times10^{-4}$ \\
\bottomrule
\end{tabular}
\end{table}

\section{Model Selection and Multi-Seed Stability}
\label{app:models}
GRU and Transformer use nested training-side selection over two fixed candidates per outer fold. GRU candidates differ in dropout/learning rate: $(0.1,10^{-3})$ versus $(0.2,3\times10^{-4})$ at fixed width 128 and two layers. Transformer candidates are two layers at $10^{-3}$ versus three layers at $3\times10^{-4}$, both width 64, four heads, FFN 128, and dropout 0.1. Every outer fold selects the first candidate. Inner train/validation sets contain 94--96 and 17 participants, respectively, with zero overlap with the 32--34 outer-test participants. JEPA remains fixed rather than searched.

\begin{table}[h]
\centering
\caption{Three-seed participant-held-out NMSE. Participant uncertainty is analyzed separately from across-seed variability.}
\small
\setlength{\tabcolsep}{5pt}
\begin{tabular}{lrrrr}
\toprule
Model & seed 0 & seed 1 & seed 2 & seed SD \\
\midrule
MLP & 0.1756 & 0.1781 & 0.1771 & 0.0013 \\
GRU & 0.1962 & 0.1979 & 0.2004 & 0.0021 \\
Transformer & 0.1764 & 0.1759 & 0.1776 & 0.0009 \\
JEPA & 0.3042 & 0.3141 & 0.3120 & 0.0052 \\
\bottomrule
\end{tabular}
\end{table}

\section{Conservative Identity Diagnostics}
\label{app:identity}
The primary identity diagnostic excludes the entire query recording from the retrieval gallery. Query-level hit@1 values are GRU 0.209, Transformer 0.189, MLP 0.156, Ridge 0.156, persistence 0.143, and JEPA 0.118 versus nominal fold-level same-patient chance 0.0086. These are descriptive; the participant-level H2 tests are not significant after correction.

As a separate sanity check on the representation pipeline, a training-side GRU linear identity probe on outer fold 0 achieves accuracy 0.395 on a capped subsample, while 100 participant-label permutations have mean accuracy 0.0067 and maximum 0.0226. This permutation test does not exclude same-recording information and is therefore \emph{not} used as evidence for cross-recording identity or as a substitute for H2; it only verifies that the probe machinery detects non-random participant structure when such information is available.

\begin{figure}[H]
\centering
\begin{subfigure}[t]{0.49\linewidth}
\centering
\includegraphics[width=\linewidth]{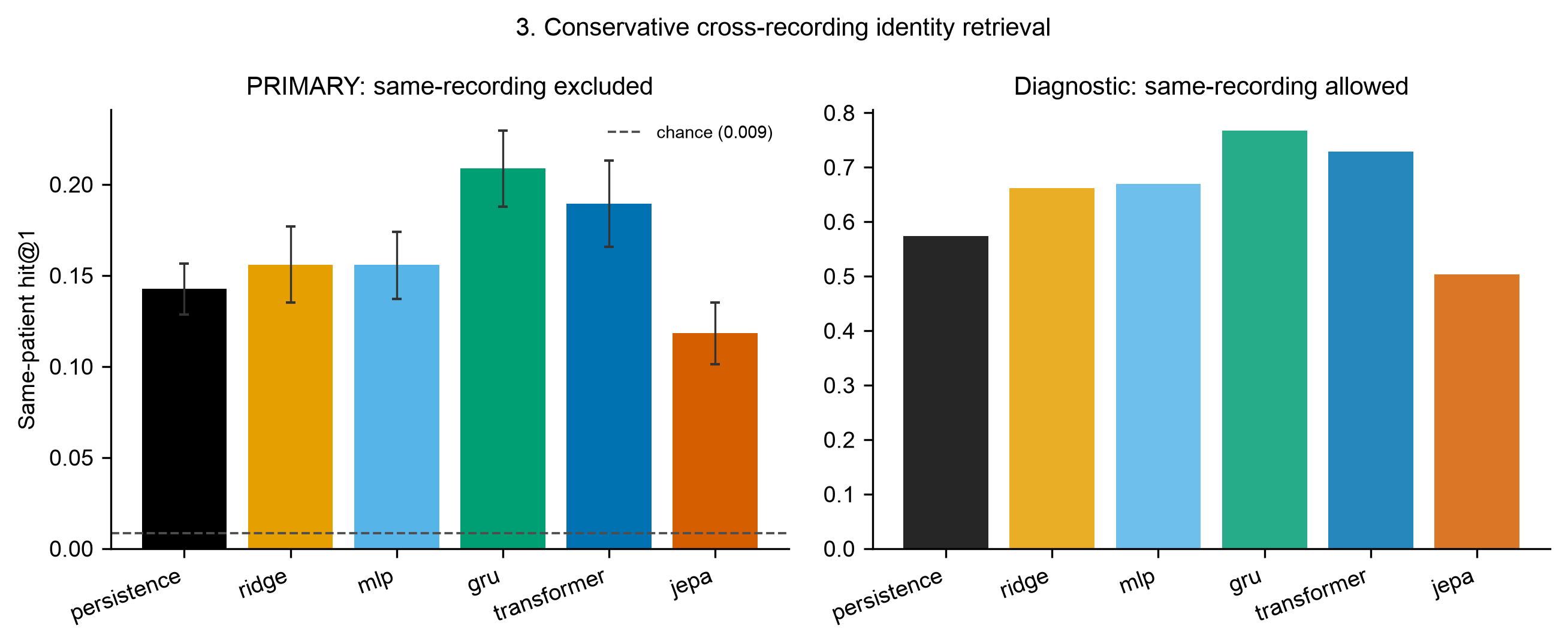}
\caption{Recording-excluded retrieval.}
\end{subfigure}\hfill
\begin{subfigure}[t]{0.49\linewidth}
\centering
\includegraphics[width=\linewidth]{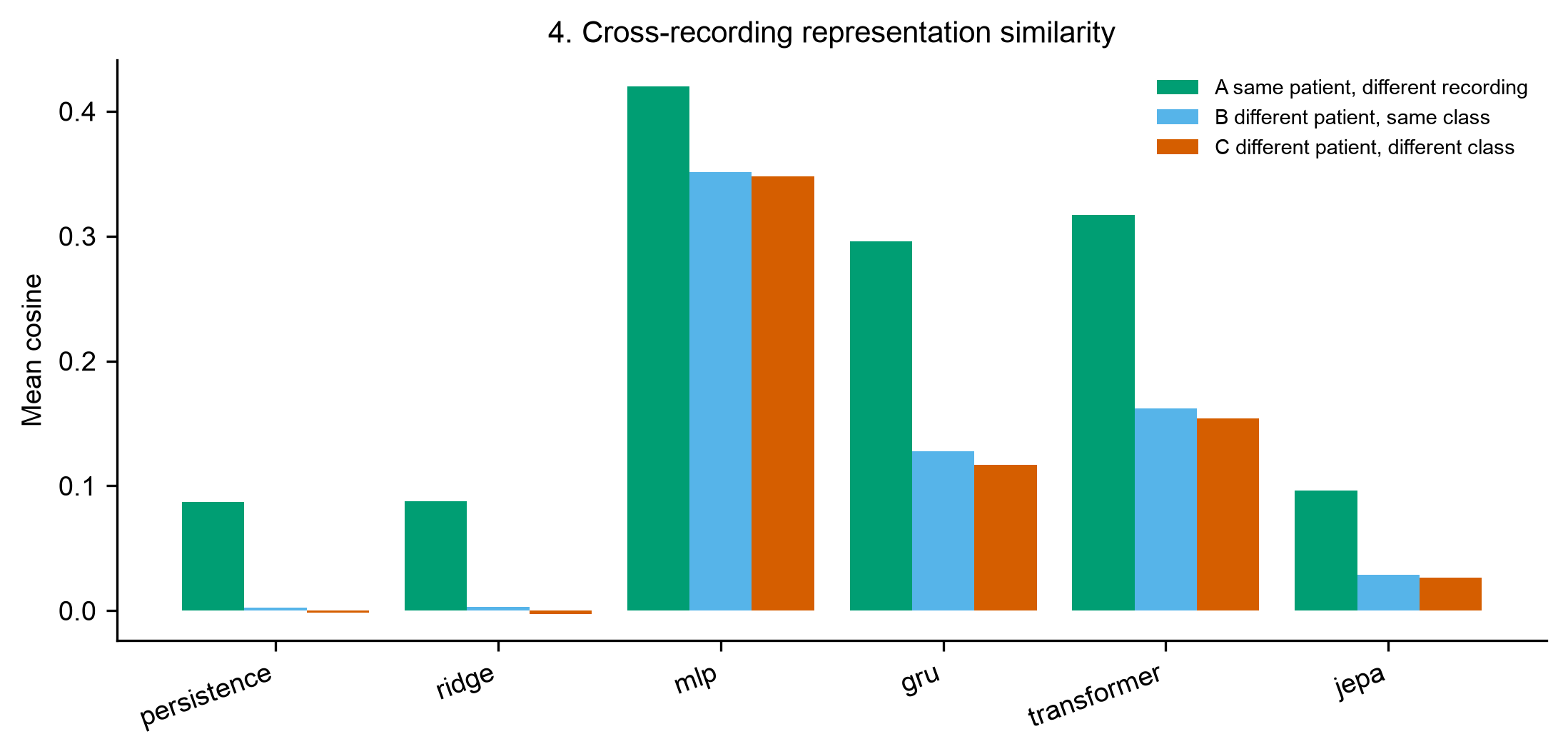}
\caption{Cross-recording cosine structure.}
\end{subfigure}
\caption{Conservative identity diagnostics. Same-recording-allowed retrieval is shown only as a diagnostic of within-walk autocorrelation; the paper's identity discussion uses recording-excluded results.}
\end{figure}

\section{Diagnosis-Stratified Descriptive Check}
\label{app:subgroups}
This analysis is exploratory and is not a disease-classification or severity analysis. Under seed-0 participant holdout, model ordering is qualitatively similar within HC and PD participants.
\begin{table}[h]
\centering
\caption{Participant-held-out NMSE by diagnosis (seed 0; participant-bootstrap 95\% CI).}
\small
\begin{tabular}{lcc}
\toprule
Model & HC ($n=72$) & PD ($n=93$) \\
\midrule
Ridge & 0.237 [0.201,0.281] & 0.252 [0.228,0.279] \\
MLP & 0.169 [0.151,0.188] & 0.181 [0.167,0.195] \\
GRU & 0.187 [0.163,0.214] & 0.203 [0.188,0.219] \\
Transformer & 0.171 [0.154,0.189] & 0.181 [0.168,0.194] \\
\bottomrule
\end{tabular}
\end{table}
No causal or clinical interpretation is assigned to between-group error differences.

\section{Additional Reliability Diagnostics}
\label{app:reliability}
\begin{figure}[H]
\centering
\begin{subfigure}[t]{0.58\linewidth}
\centering
\includegraphics[width=\linewidth]{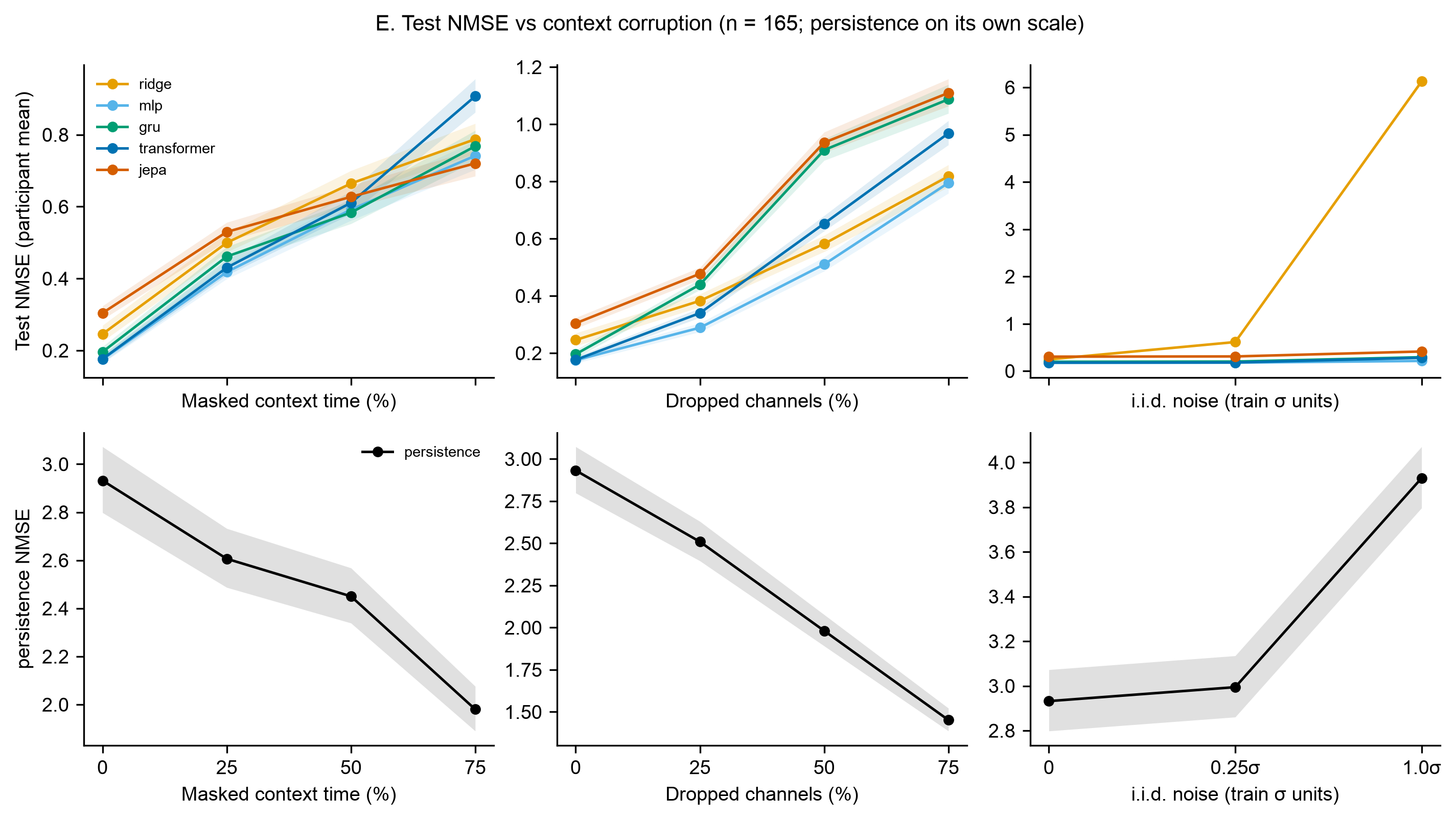}
\caption{Full partial-observation curves.}
\end{subfigure}\hfill
\begin{subfigure}[t]{0.40\linewidth}
\centering
\includegraphics[width=\linewidth]{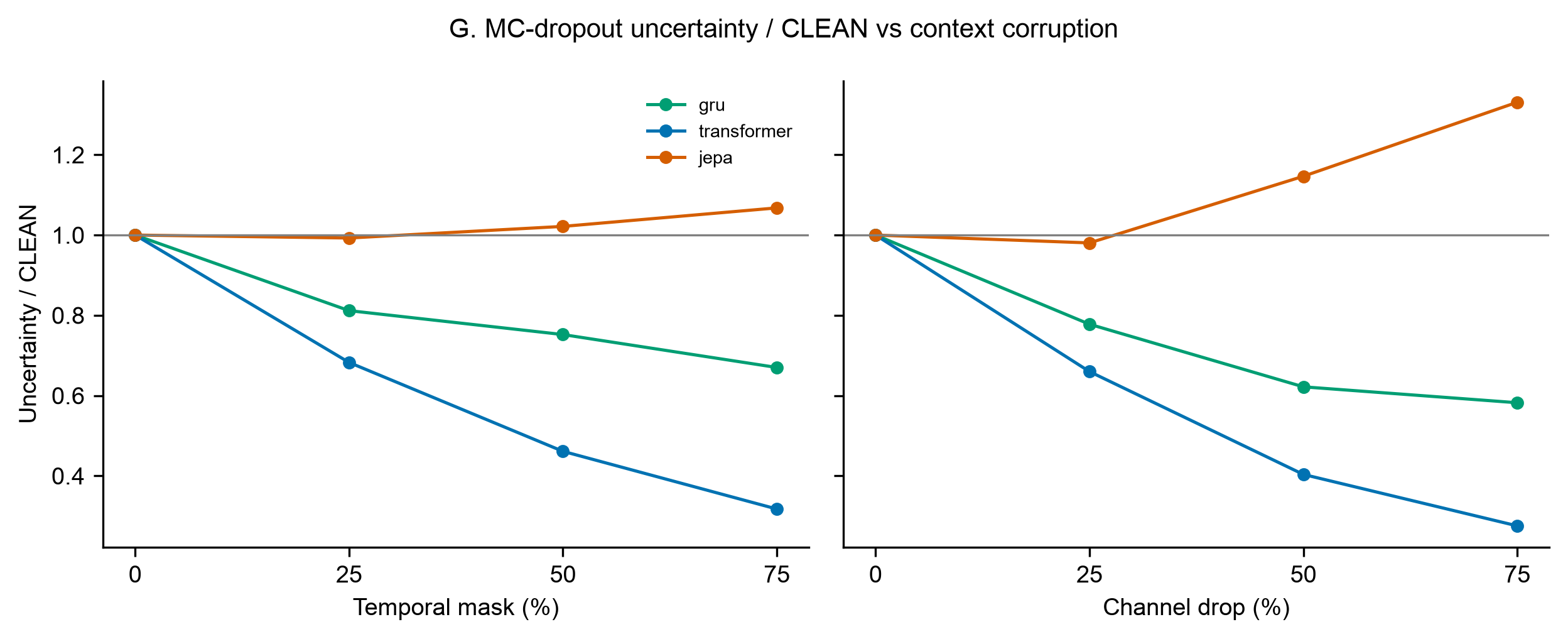}
\caption{Uncertainty across severity.}
\end{subfigure}
\caption{Full severity sweeps used by the partial-observation and reliability analyses.}
\label{fig:fullrobust}
\end{figure}

The full corruption-severity curves are shown in Figure~\ref{fig:fullrobust}. They include 25/50/75\% temporal masks and channel removal plus low/medium Gaussian noise; the main paper displays the 50\% conditions for legibility.

\begin{figure}[H]
\centering
\includegraphics[width=0.82\linewidth]{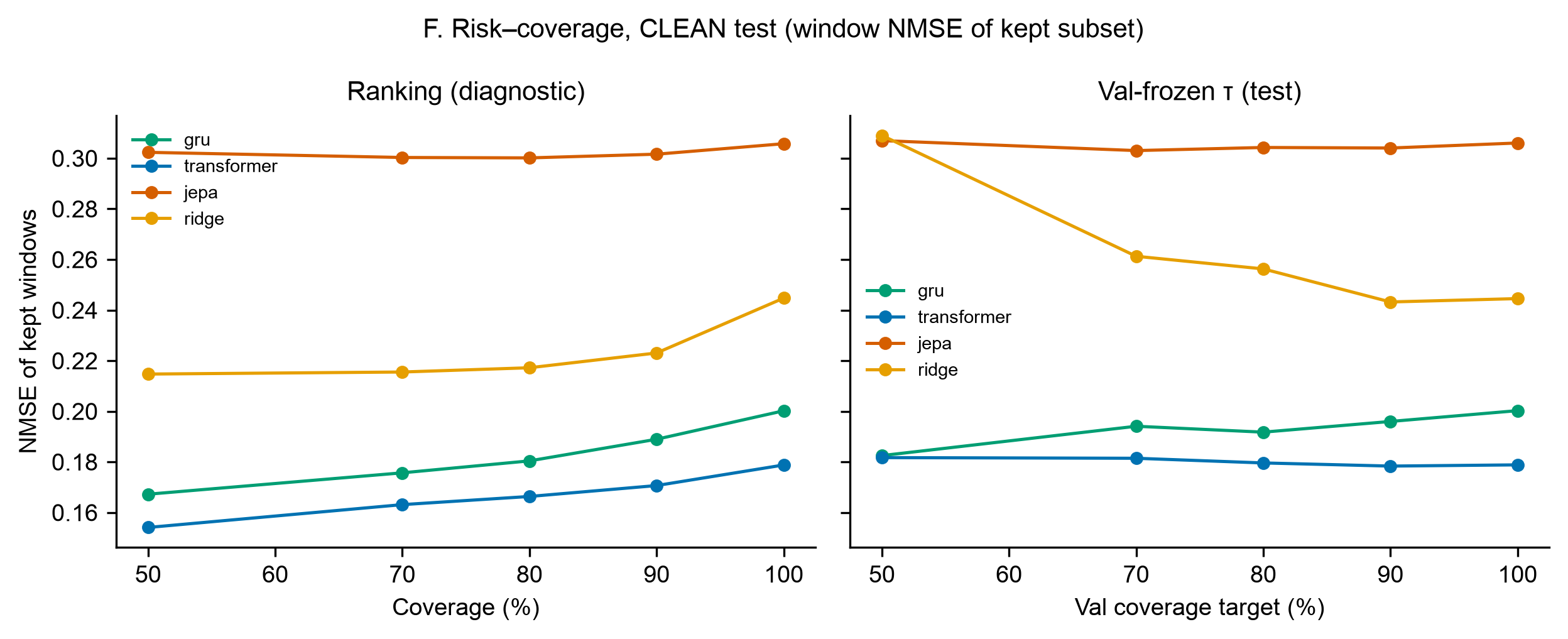}
\caption{Risk--coverage diagnostics on clean participant-held-out test data. Ranking (left) is diagnostic; validation-frozen target-coverage thresholds (right) are the deployable-style analysis.}
\label{fig:risk}
\end{figure}

Figure~\ref{fig:risk} distinguishes two selective-prediction views. The left panel sorts test windows by observed uncertainty to trace a diagnostic risk--coverage curve; it does not define a deployable threshold. The right panel applies thresholds selected from training-side validation quantiles. Only GRU shows a modest improvement near 50\% target coverage; Transformer and JEPA are flat or slightly worse.

\begin{figure}[H]
\centering
\includegraphics[width=0.96\linewidth]{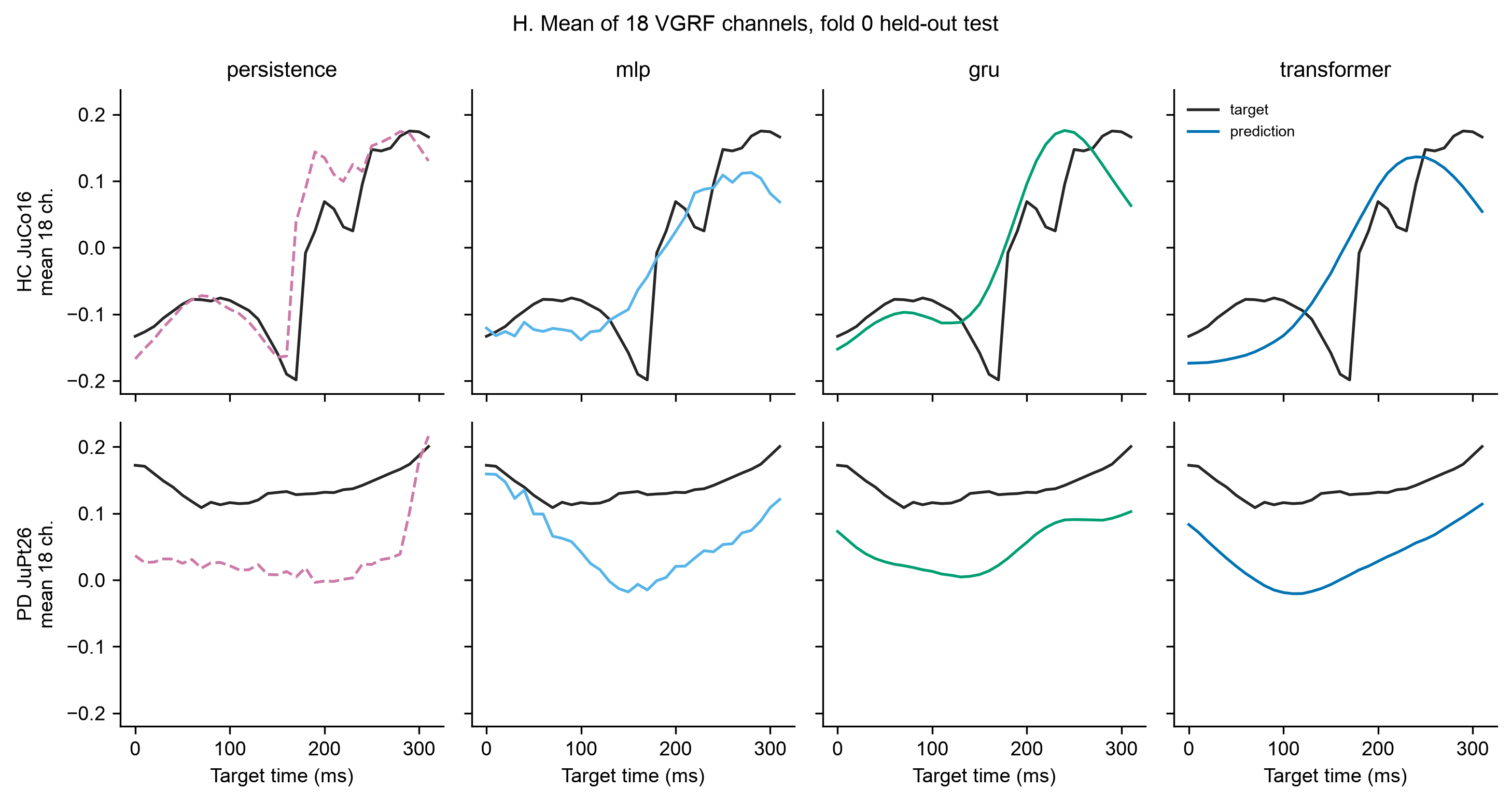}
\caption{Representative mean and per-channel future VGRF targets and predictions from a held-out fold.}
\label{fig:segments}
\end{figure}

Representative predictions in Figure~\ref{fig:segments} are illustrative only and were not selected as best cases.

\section{Reproducibility and Artifact Integrity}
The final research handoff contains fixed fold manifests, model configurations, seed manifests, participant/recording/window-level result tables, statistical-test tables, claim-to-evidence provenance, and raw figure sources. The complete repository test suite reports 106 passing tests, including participant-overlap, nested-selection, blocked-overlap, LOSO, cross-recording, and result-tracing checks. The portability audit passed the full test suite and path/manifest checks in the declared environment; because local disk space was below the project's safety floor, it did not create a completely fresh virtual environment, and we disclose that limitation rather than calling the check fully clean-room. Experiments ran locally on Apple Silicon using PyTorch MPS with CPU fallback; MPS is not claimed to be bitwise deterministic. The OSF artifact provides the code and paper-facing outputs without redistributing the raw PhysioNet data; large checkpoints may be omitted when the checkpoint manifest preserves model/fold/seed/configuration and checksums.

\end{document}